\documentclass[10pt, conference, compsocconf]{IEEEtran}

\pdfoutput=1
\usepackage[numbers]{natbib}
\usepackage{gensymb}
\usepackage{booktabs}
\usepackage{graphicx}
\usepackage[hidelinks]{hyperref}
\setkeys{Gin}{width=\linewidth,totalheight=\textheight,keepaspectratio}
\graphicspath{{graphics/}} 
\usepackage{amsmath}  
\usepackage{multicol}
\usepackage{caption}
\usepackage{subcaption}
\usepackage{color,soul}

\title{Towards developing a realistic robotics simulation environment of an indoor vegetable greenhouse}

\author{\IEEEauthorblockN{Brent Van De Walker, Brendan Byrne, Joshua Near, \\ Blake Purdie, Matthew Whatman, David Weales, Cole Tarry, Medhat Moussa}
\IEEEauthorblockA{School of Engineering\\
	University of Guelph\\
	Guelph, Canada\\
	bvandewa, dweales, mmoussa, ctarry, @uoguelph.ca}
}

\begin{document}
	
\maketitle

\section{Introduction}

Greenhouses are responsible for a significant amount of modern crop production worldwide. In Canada, the vegetable greenhouse industry was responsible for almost five hundred million kilograms of tomato harvest in 2020 alone \cite{government_of_canada_area_2021}, resulting in a farm gate value of \$116 million. Historically in Canada, there has been a greenhouse labour shortage, as there were 2800 unfilled jobs in 2014, costing the horticultural industry approximately \$100 million \cite{reuters_canadas_2019}. Automating the harvesting process can reduce labour cost and eliminate shortages. Given the significant clutter and occlusion typically found in a vegetable greenhouse, an important step is to create a realistic simulation environment to enable testing of various techniques without the need for expensive experimental setups or slow down production in a commercial greenhouse. However, given the complexity of a real greenhouse environment, it is difficult to build a simulation that has a realistic reconstruction of that environment. In this paper, we present a approach for developing simulated environment from captured 3D images of a commercial greenhouse. The 3D data are imported into a robotics simulator and tuned to enable testing of various harvesting operations. We present one scenario where a robot is attempting to pick a tomato on a vine. Various path planning approaches can be tested and optimized before real-world testing. 

Although the clutter and plant size varies greatly throughout the season, tomatoes aren't typically picked for approximately six months after planting, when they have grown considerably \cite{calpas_commercial_2003}. Also, the plants are typically de-leafed before harvesting, so it is reasonable to assume less clutter. There have been several attempts to develop a robotics simulation environment for various agriculture projects including greenhouses. The CROPS (Clever Robots for Crops) project \cite{best_crops_2015} used a simulation in Gazebo to evaluate the robot without the requirement of taking the actual robot into the field. Nguyen et al. developed a framework of the motion planning task for the CROPS robot \cite{nguyen_task_2013}. He use an apple tree simulation environment with only the tree trunk and branches considered as obstacles. Shamshri et al.  \cite{shamshiri_robotic_2018} provide a review of various simulation software that can be used for developing simulation environment for agriculture applications including the one used in this paper V-rep. In all of the previous efforts, artificial model of plants were used in the simulation. While these models can be adequate for some operations, in this paper we attempt to bridge the gap between the real-world environment and the simulated one by including data captured directly from a commercial greenhouse. This pipeline will enable not just a snap shot of the real-world environment but also allows accurate capturing of the environment clutter and how it changes as data is updated on a regular basis.

\section{Materials and methods}
 
\subsection{Robotics Simulation Platform}

There are multiple different simulators and virtual environments that can be used for testing designs. Each simulator has its own set of merits and limitations which may cause one type of simulator to be more suited to a specific task than the others. A comparative paper which addresses this issue can be seen in \cite{shamshiri_simulation_2018}, where simulators  VREP, Gazebo, and ARGoS are compared against eachother based on their respective performances within similar scenes. Small and large scenes were compared with a changing number of robots. Of the three simulations, VREP achieved the slowest simulation speed and the highest memory usage. ARGoS had the fastest simulation speed for small scenes, but was taken overby Gazebo as the scene size increased. However, despite V-REP's lacking simulation speed, carefully setting simulation parameters and optimizing 3D models can drastically increase its performance. This is also shown in \cite{pitonakova_feature_2018}, where a feature comparison is done between the V-REP, Gazebo, and ARGoS simulators. With respect to the features available in the simulators, V-REP has the largest collection of features such as a scene editor, 3D model importing, and mesh manipulation. A significant disadvantage of V-REP is its inability to define a scene in an XML file. This would allow multiple experiments with varying parameter values to be run automatically. While V-REP is the most complex of the three simulators, it offers many features which are not available otherwise.

\subsection{3D mapping of a vegetable greenhouse}

In order to adequately create a simulation which mimics a greenhouse environment, the environment should reflect actual data from a greenhouse. 

Figure \ref{fig:greenhouseenv} below shows an image taken from a commercial tomato greenhouse in Ontario, Canada. This represents a sample of the environment we wish to mimic in the simulation environment.

\begin{figure}[h]
	\centering
	\subfloat{\includegraphics[width=0.7\linewidth]{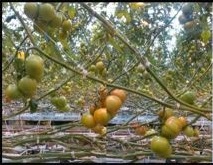}}
	\caption{An image from a commercial tomato greenhouse}
	\label{fig:greenhouseenv}
\end{figure}

Since the objective of the simulation environment is to enable simulating of a robotic picking operations, it is important to have a 3D map of the environment. To accomplish this task, the images were collected using a three-camera system mounted on a tripod placed approximately 50 cm from the nearest plants. The tripod was manually moved by a human along the greenhouse rows and images were collected approximately every 10 cm. This process was repeated for two greenhouse rows. 

\subsubsection{RGB-D image capture system}
\begin{figure}[h]
	\centering
	\includegraphics[width=0.9\linewidth]{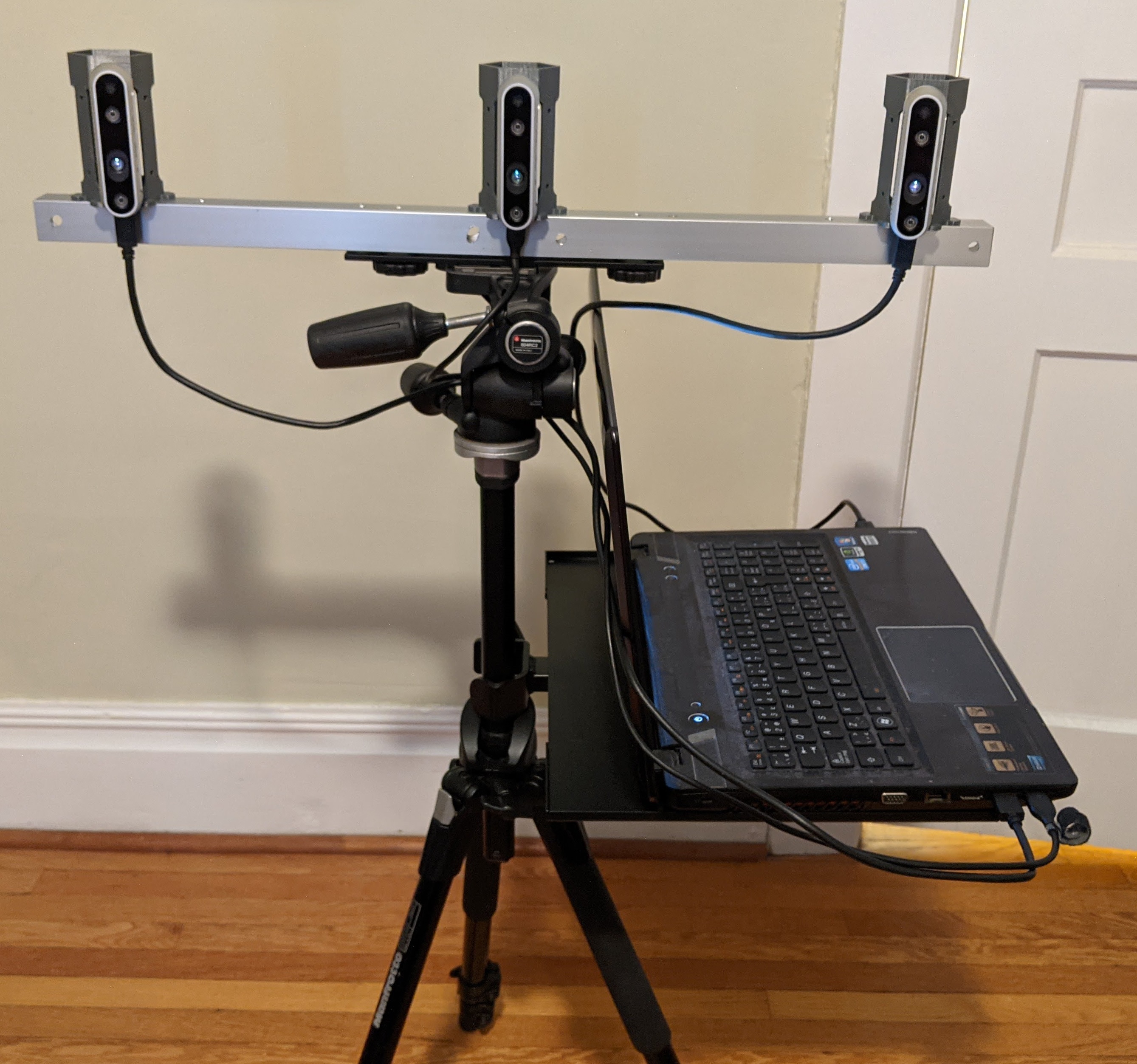}
	\caption{Setup}
	\label{fig:cameramount}
\end{figure}

The RGB-D image capture system included three Intel® RealSense™ D435 cameras mounted on a tripod along with a laptop for collecting and storing the images. The cameras were placed on a metal beam 25 cm apart as shown in Figure \ref{fig:cameramount}. The outer two cameras were angled 25\degree toward the center in order to get a more complete 3D view of the plants.

\subsubsection{Image processing}

The first step in the algorithm is to identify the matching regions in consecutive RGB images. The RGB processing is performed using the images from the center camera as it stays parallel to the row as the tripod moves.

\begin{figure}[h]
	\centering
	\subfloat[First image]{%
		\includegraphics[width=0.9\linewidth]{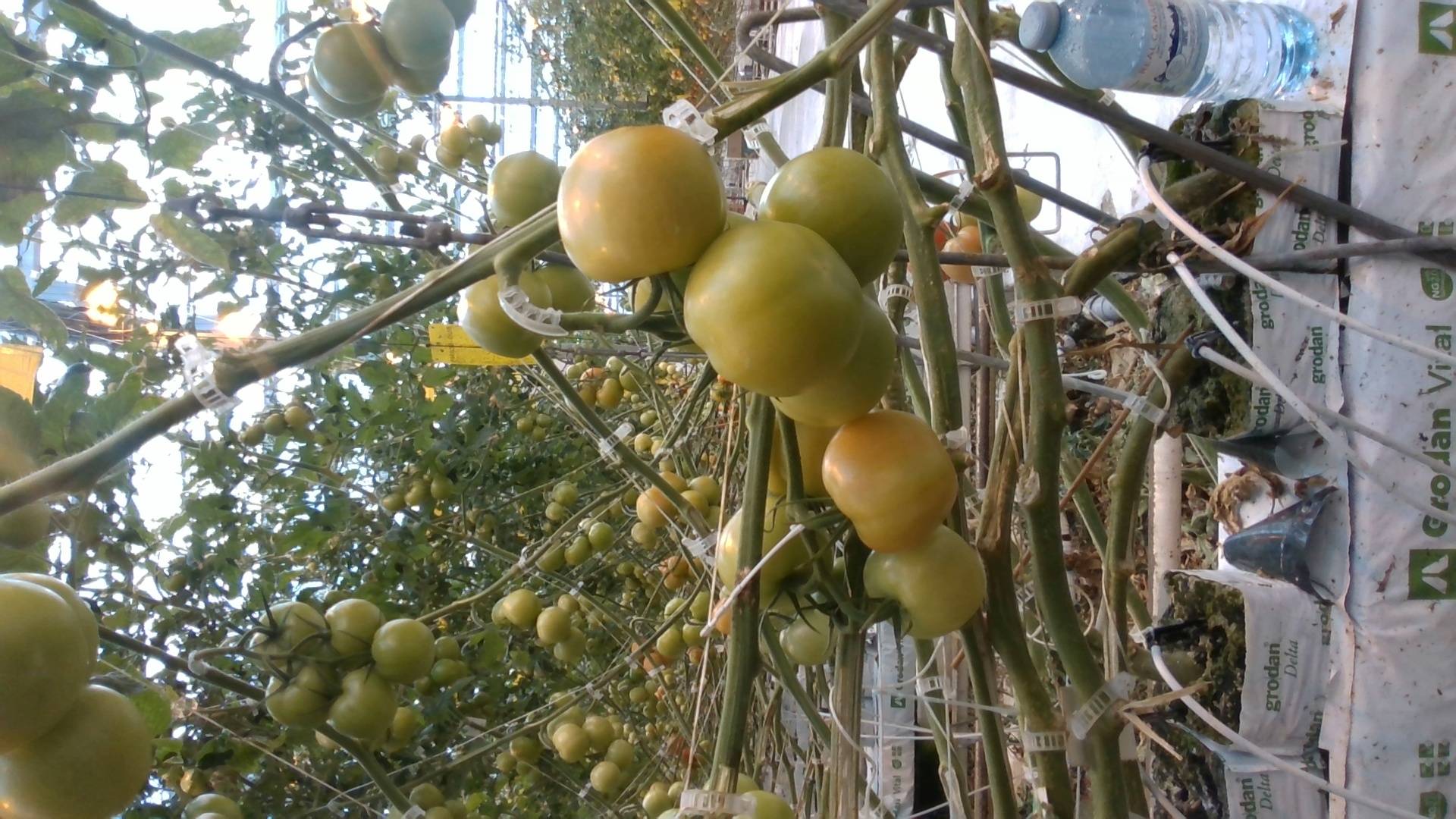}}
	\hfill
	\subfloat[Second image]{%
		\includegraphics[width=0.9\linewidth]{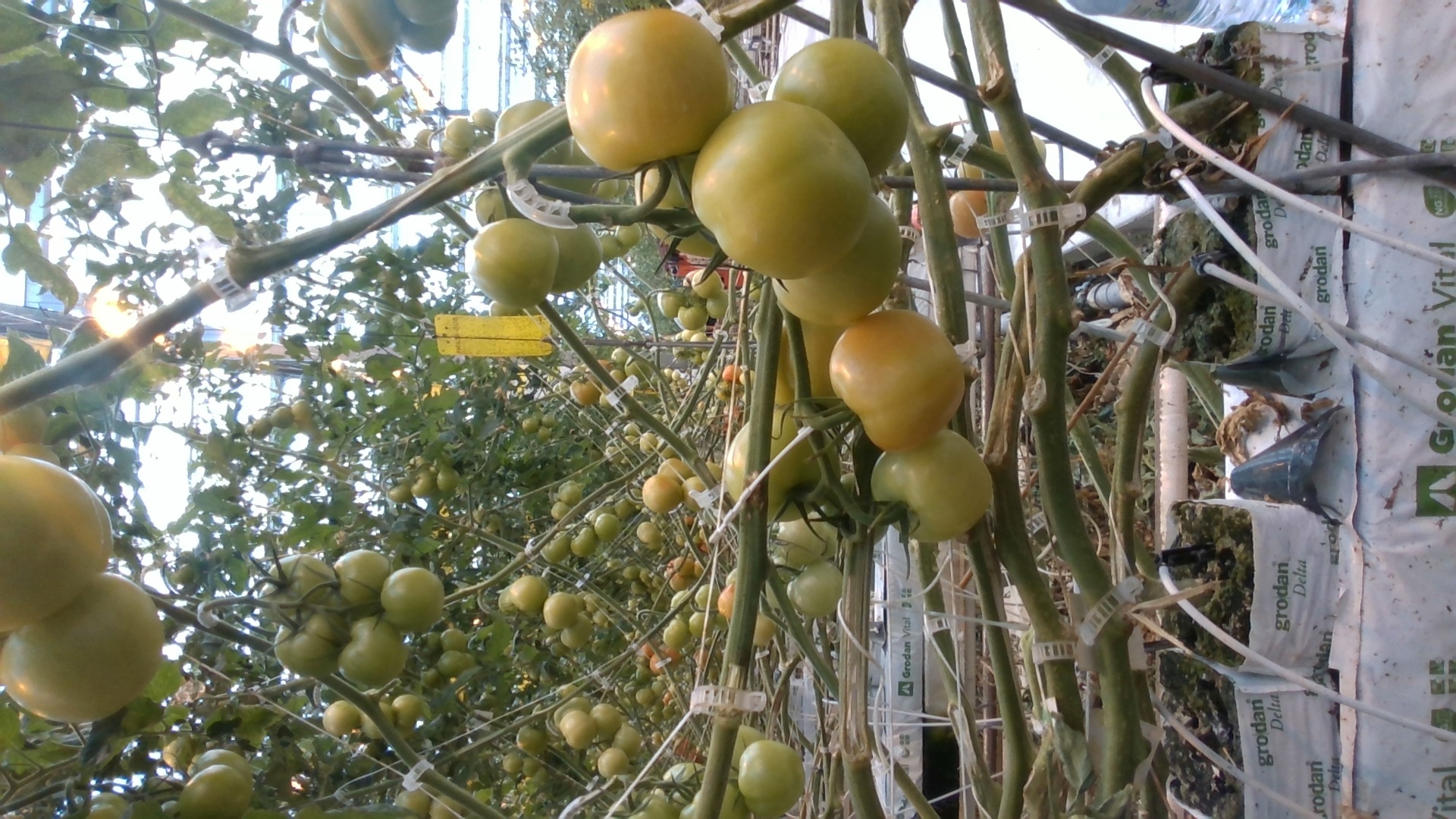}}
	\caption{Two images of the row taken approximately 10 cm apart.}
	\label{fig:rowimgs}
\end{figure}

\begin{figure}[h]
	\centering
	\subfloat[First image]{%
		\includegraphics[width=0.9\linewidth]{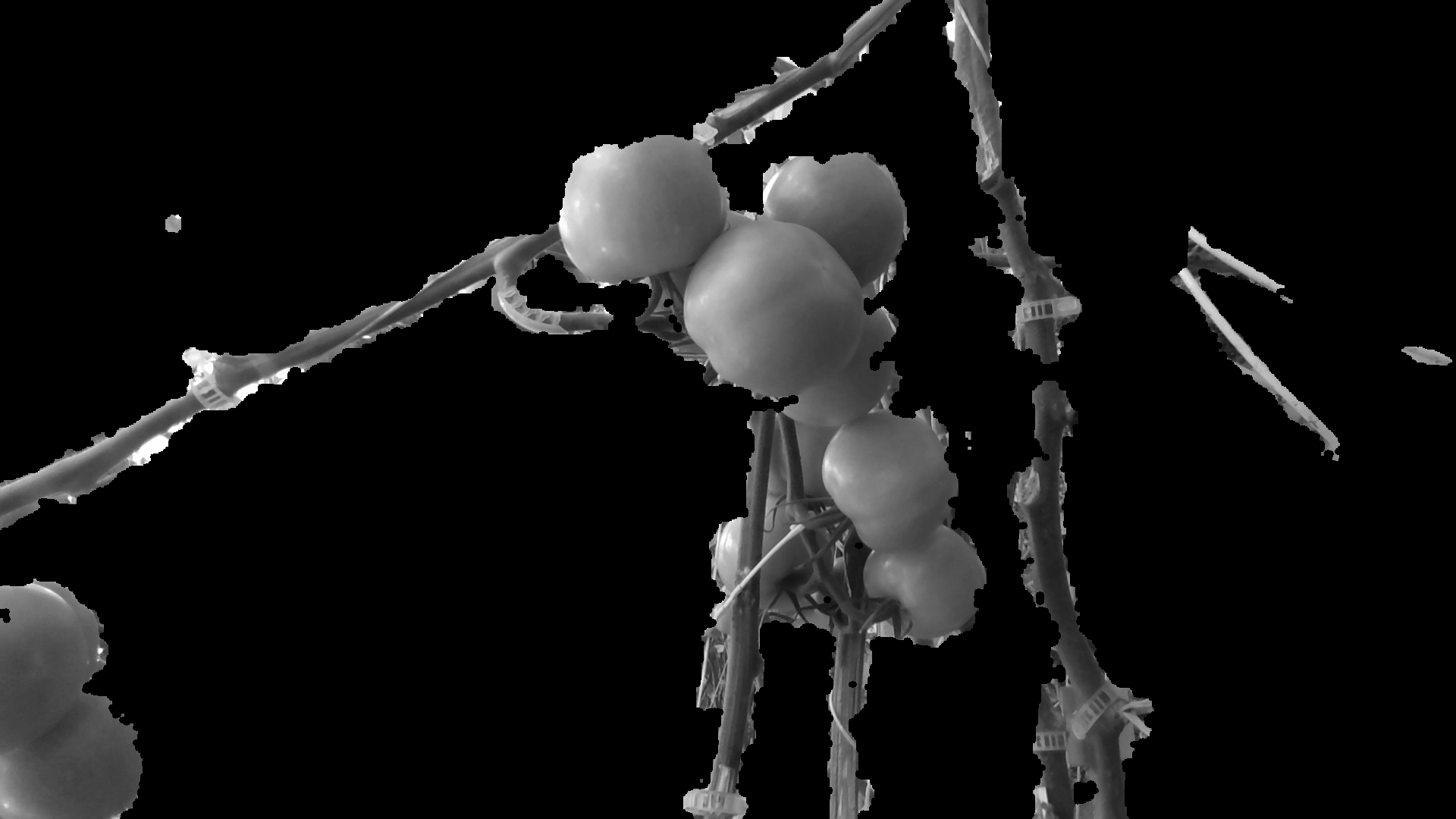}}
	\hfill
	\subfloat[Second image]{%
		\includegraphics[width=0.9\linewidth]{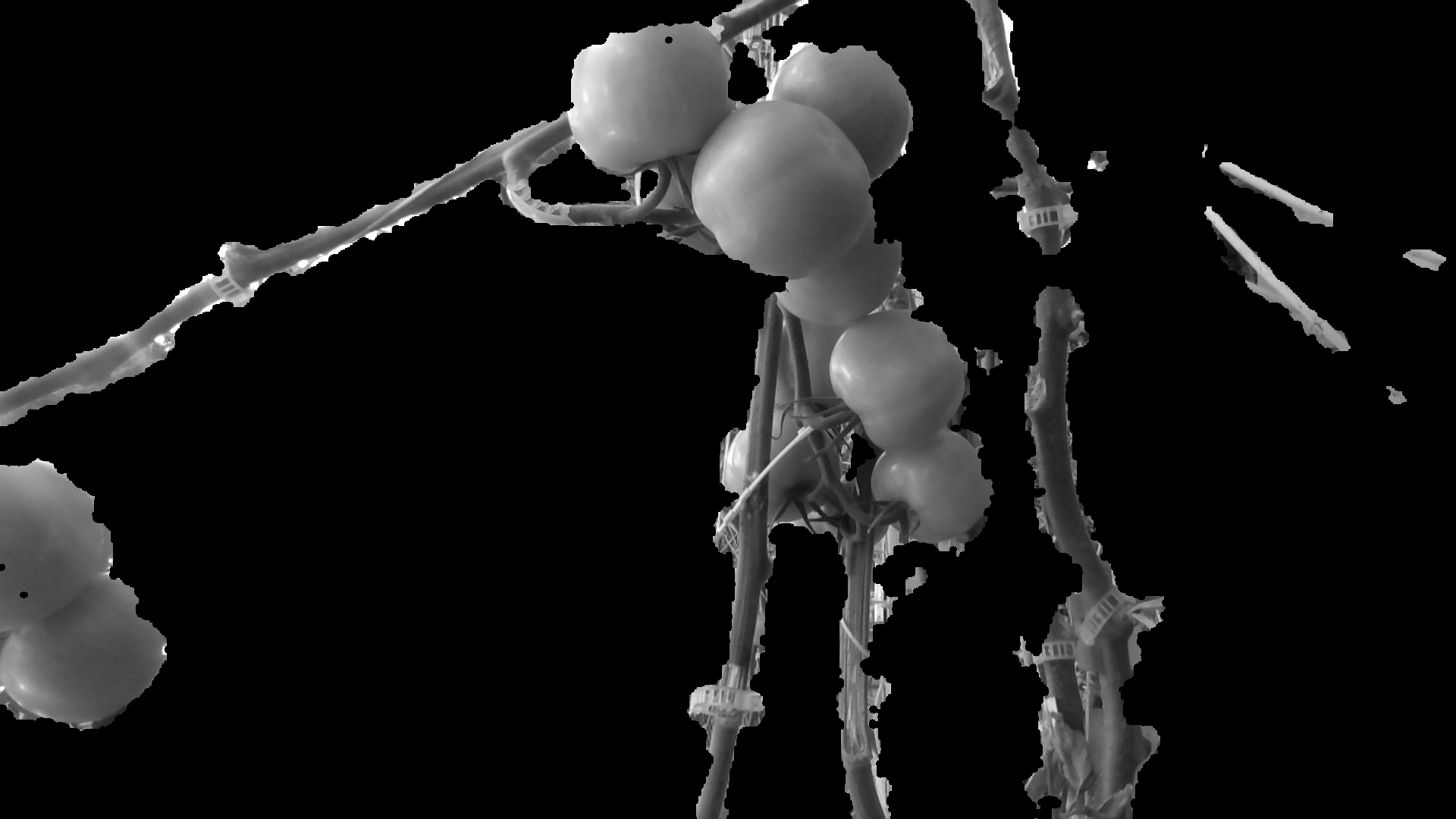}}
	\caption{Two images from different locations with areas further than 60 cm from the camera removed.}
	\label{fig:roi}
\end{figure}

To begin, we remove the background plants by identifying regions in the depth image that are more than 60 cm from the camera. By blacking out the background, we are able to focus on the nearest vines and tomatoes. A sample of two successive images with the background removed is shown in Figure \ref{fig:roi}.

\begin{figure}[h]
	\centering
	\subfloat[First image]{%
		\includegraphics[width=0.49\linewidth]{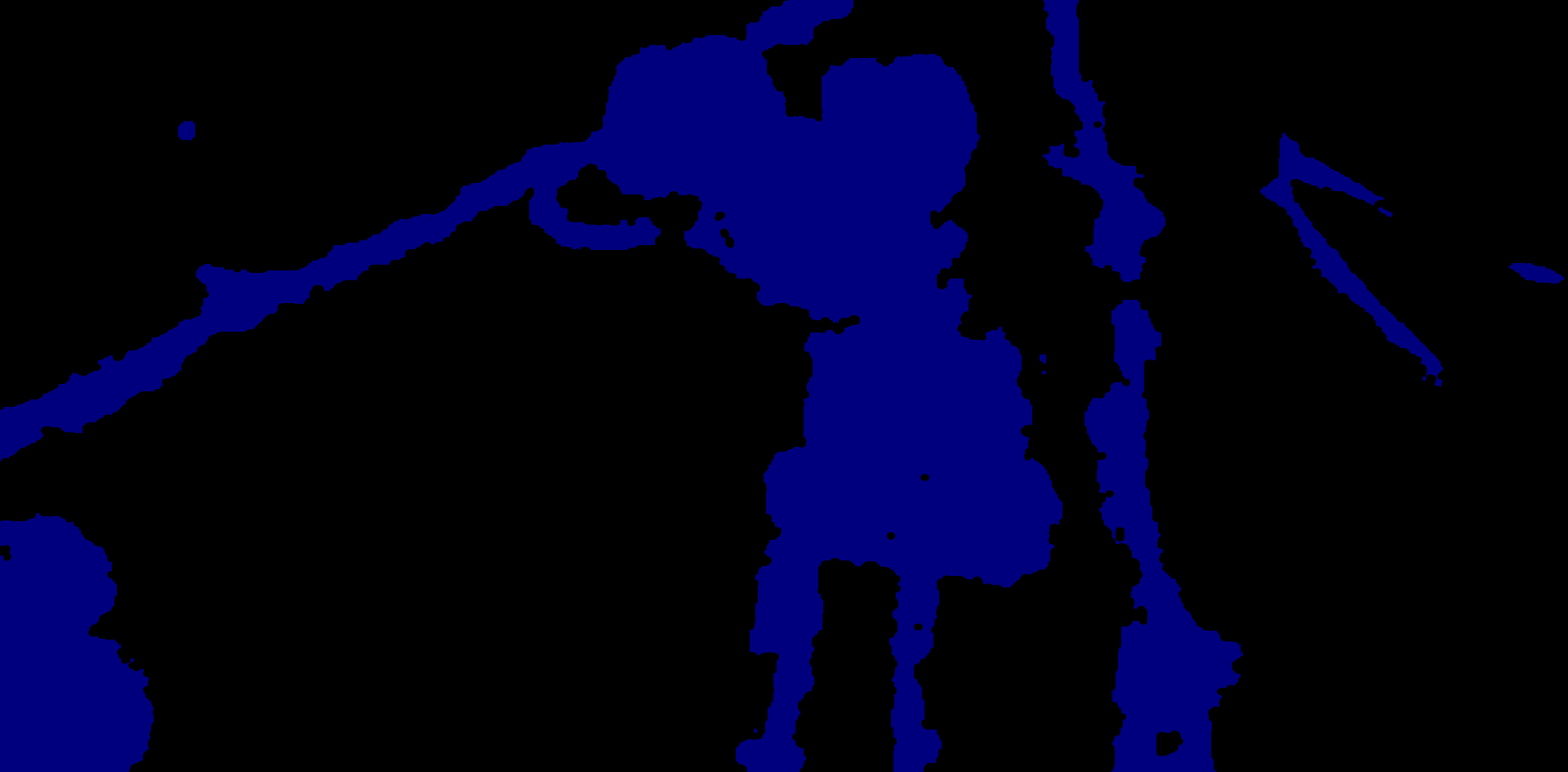}}
	\hfill
	\subfloat[Second image]{%
		\includegraphics[width=0.49\linewidth]{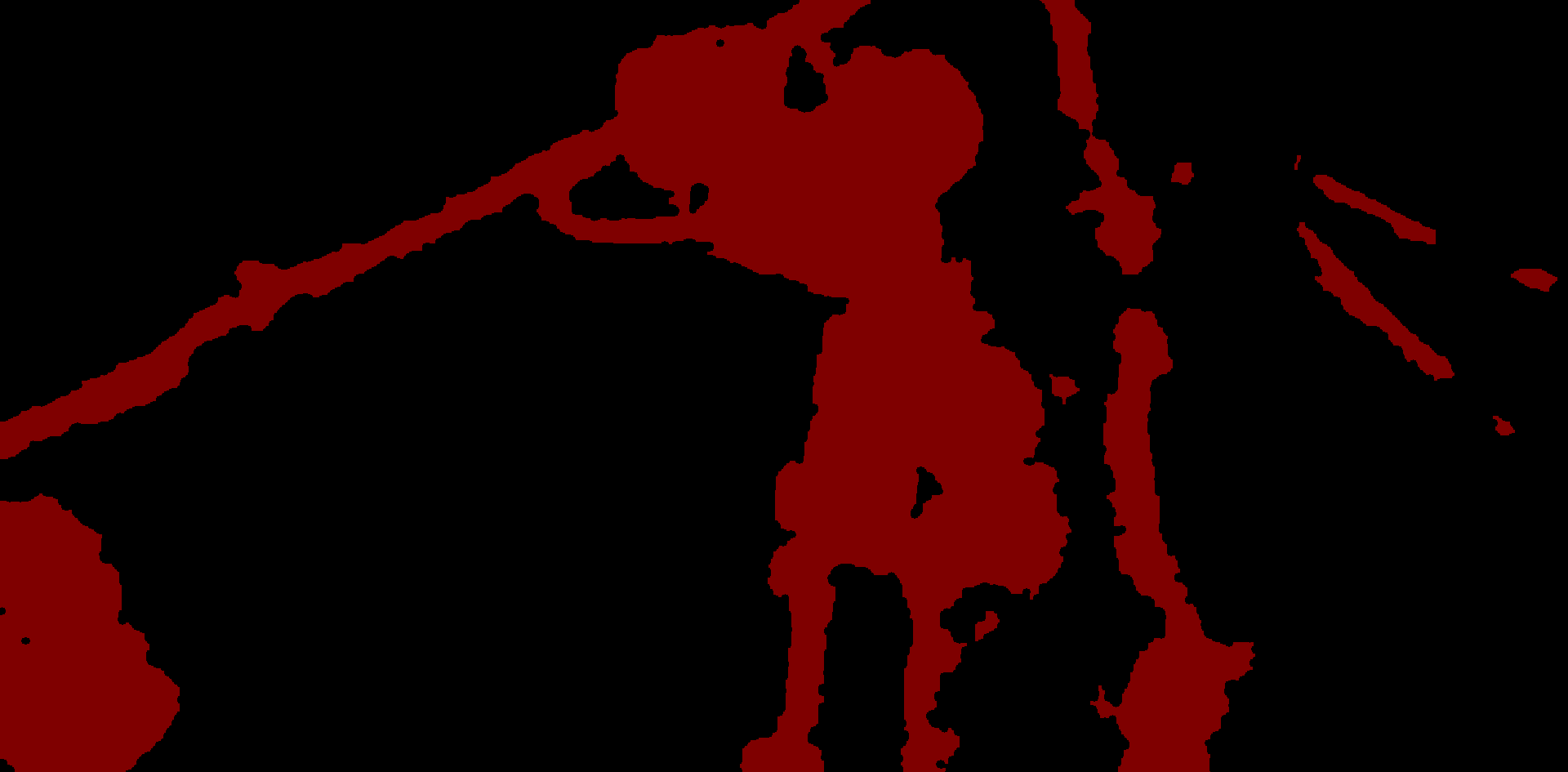}}
	\hfill
	\subfloat[Both images overlaid]{%
		\includegraphics[width=\linewidth]{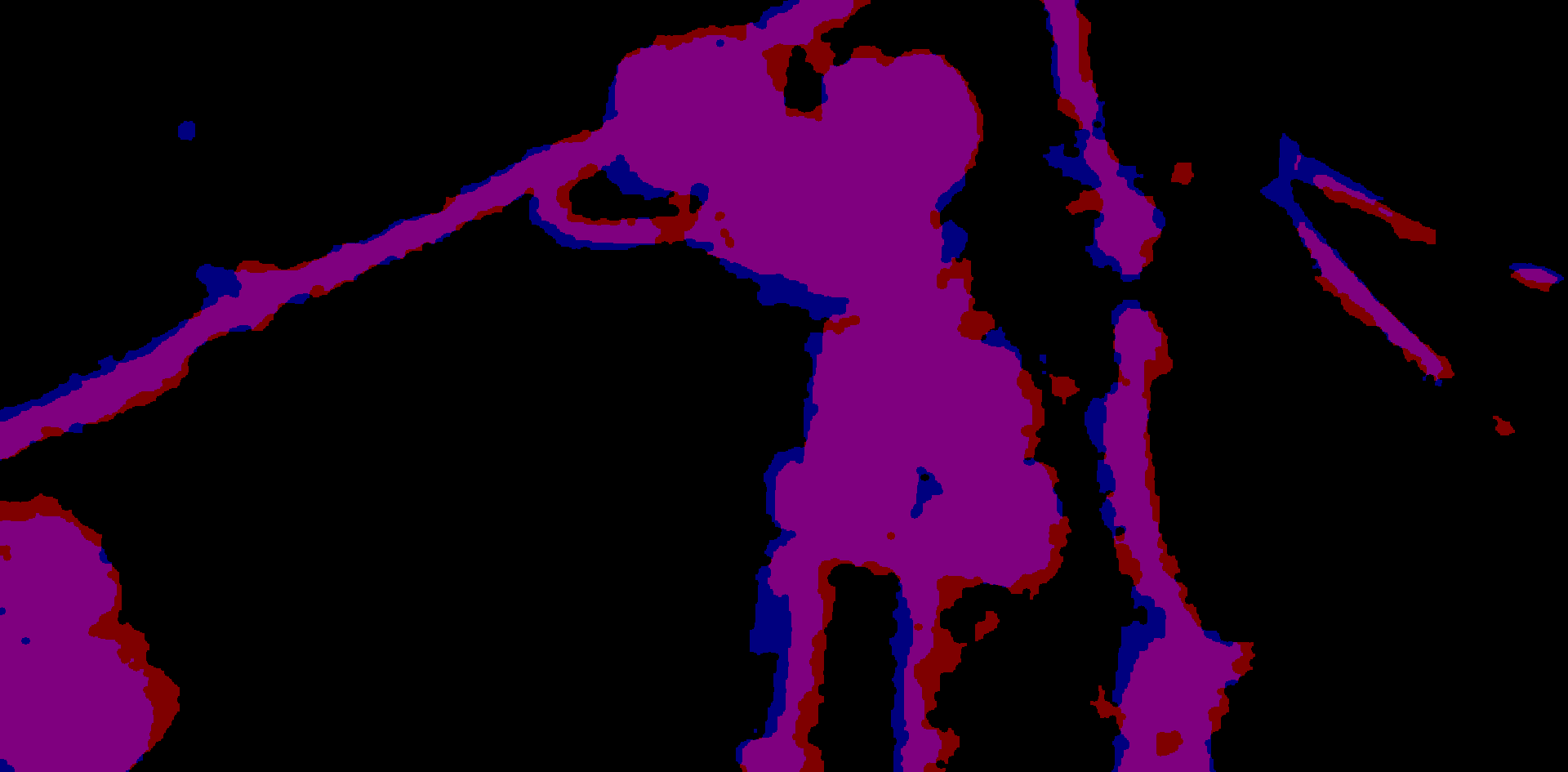}}
	\caption{Two images from different locations shifted to maximize overlapping areas. The pink regions represent the overlap.}
	\label{fig:roicompare}
\end{figure}

Next, we shift one of the images to attempt to match the foreground with the other image. As the cameras move parallel to the row, the features in the foreground in the successive images will be a number of pixels higher than the previous images. We will attempt to match based only on the general shape of the foreground by setting the entire foreground white and the background black. Next, we gradually shift the first image up a few pixels at a time and subtract the images from one another until we find the pixel difference that produces the least difference between the images. An example of the comparison after shifting the first image is shown in Figure \ref{fig:roicompare}, where the blue and red images combine to show the shared purple regions between the images.

\subsubsection{3D processing}
Once the shared regions in the successive images have been identified, we further refine the registration using the 3D. First we examine all of the shared pixels we identified from the RGB and compare the XYZ values to identify the average distance between each shared pixel in the images. While we know the tripod was manually moved approximately 10 cm along the row for each image, this provides us with a more accurate distance the cameras were moved.

\begin{figure}[h]
	\centering
	\subfloat[Before ICP]{%
		\includegraphics[width=0.9\linewidth]{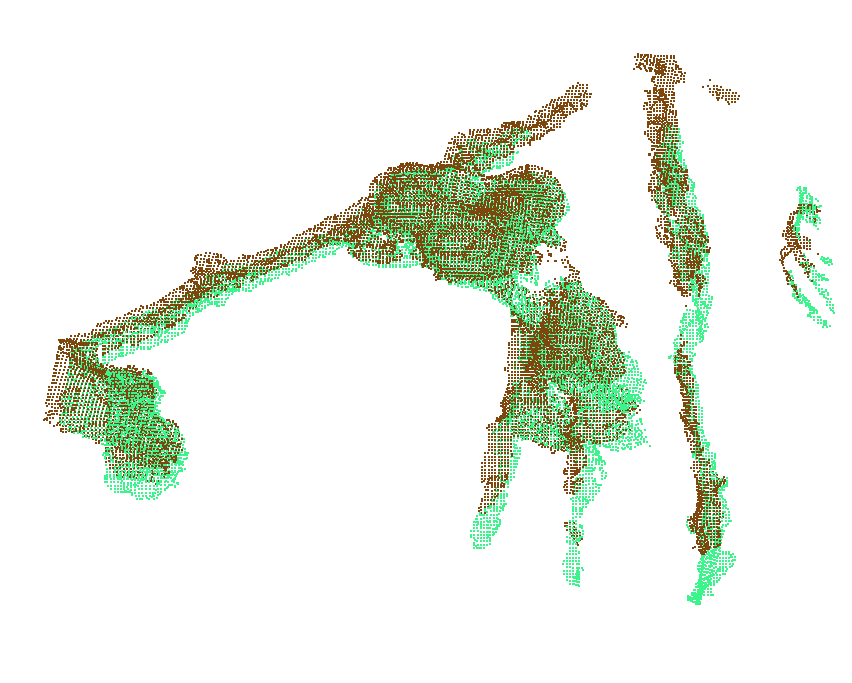}}
	\hfill
	\subfloat[After ICP]{%
		\includegraphics[width=0.9\linewidth]{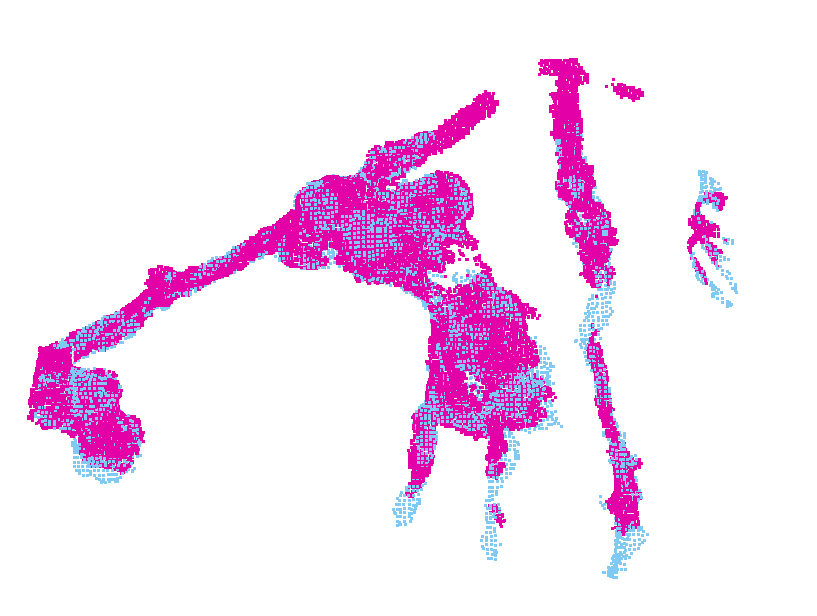}}
	\caption{Images of the two point clouds before applying ICP and after applying ICP.}
	\label{fig:icp}
\end{figure}

Now that we know the distance we can shift the 3D point clouds so that they are roughly aligned. As there may have also been some small changes in the angle of the cameras or distance from the plants, we can further refine the registration of the two point clouds using iterative closest point (ICP). We use ICP implemented in the Python Open3D library to do so. An image of the two point clouds before and after using ICP is shown in Figure \ref{fig:icp}.   

Now that we know the distance we can shift the 3D point clouds so that they are roughly aligned. As there may have also been some small changes in the angle of the cameras or distance from the plants, we can further refine the registration of the two point clouds using iterative closest point (ICP). We use ICP implemented in the Python Open3D library to do so. An image of the two point clouds before an after using ICP is shown in Figure \ref{fig:icp}.

The RGB and 3D processing stages are repeated for all the available images to create a complete 3D point cloud of the row. In some cases, the RGB or 3D processing may fail to properly register two successive images together. In this case, the second image is skipped over and we attempt to register the third image with the first image in the succession.

\begin{figure}[h]
	\centering
	\subfloat[Cloud of 50 successive images displayed in different colours]{%
		\includegraphics[width=0.9\linewidth]{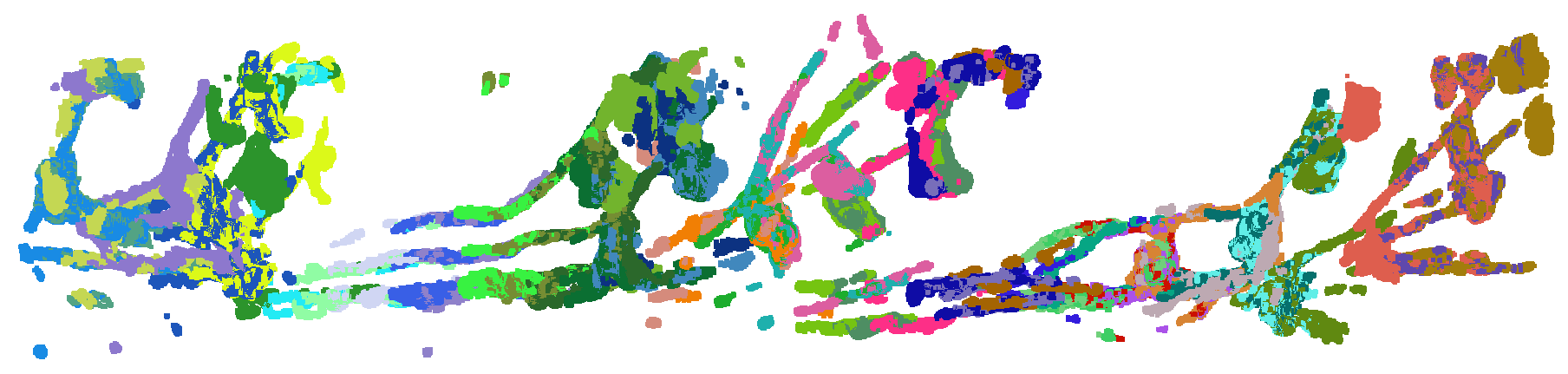}}
	\hfill
	\subfloat[Cloud after outlier removal]{%
		\includegraphics[width=0.9\linewidth]{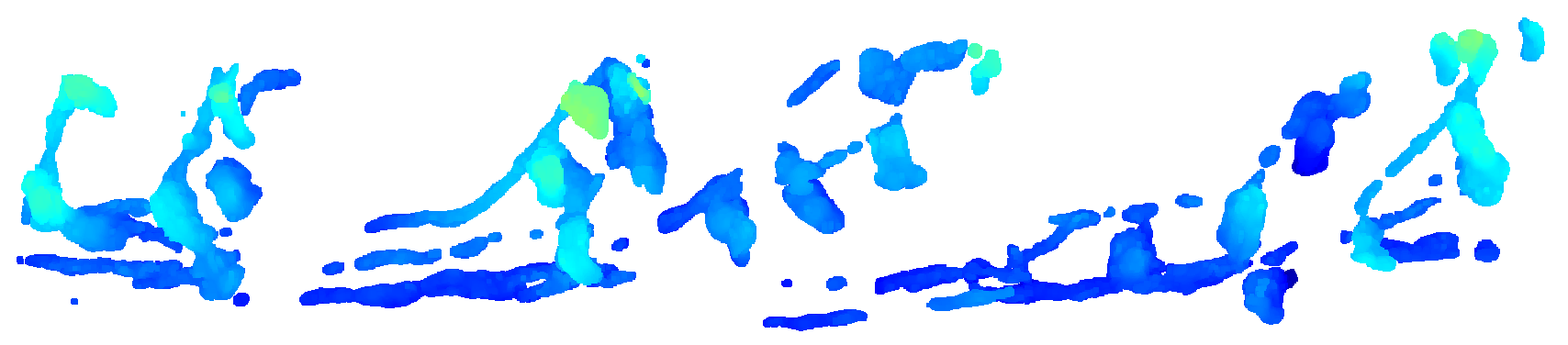}}
	\caption{Images a point cloud using 50 successive images of the greenhouse row before and after applying outlier removal.}
	\label{fig:outlier}
\end{figure}

Once all of the successive images of the greenhouse row have been combined, some postprocessing is performed to simplify the point cloud. We use a radius outlier removal to remove any points that do not have a high degree of overlap. This removes the data from any remaining point clouds that failed to properly register in the RGB processing and ICP stages. A sample point cloud using 50 successive images before and after outlier removal is shown in Figure \ref{fig:outlier}.

\subsection{Simulation environment}

The first step was to convert the environment point cloud into a triangular mesh. Once the mesh was created it was then imported into CoppeliaSim. After the mesh was imported it was then decomposed into cylinders and spheres, representing branches and tomatoes. The position of the branches and tomatoes were determined using the additionally provided images of the row of tomatoes. 

The point cloud was originally composed of 295,490 points; at this density it was very computationally taxing to convert the point cloud into a triangular mesh. In addition to it taking over an hour to convert the point cloud into a mesh, it was effectively impossible to use within CoppeliaSim due to the complexity of the mesh, having 398,571 faces. Decimation of the mesh within CoppeliaSim was attempted but found to be unable to sufficiently simplify the mesh, only being able to bring the mesh down to 143,422 faces. To obtain a less complex mesh the mesh creation process was repeated with a simplified point cloud. The point cloud was simplified to a target of 10,000 points, resulting in 10,346 points. Once the complexity of the point cloud was reduced, the point  cloud was again converted into a triangular mesh. The final resulting mesh had 17,967 faces.

\begin{figure}[h]
	\includegraphics[width=0.5\textwidth]{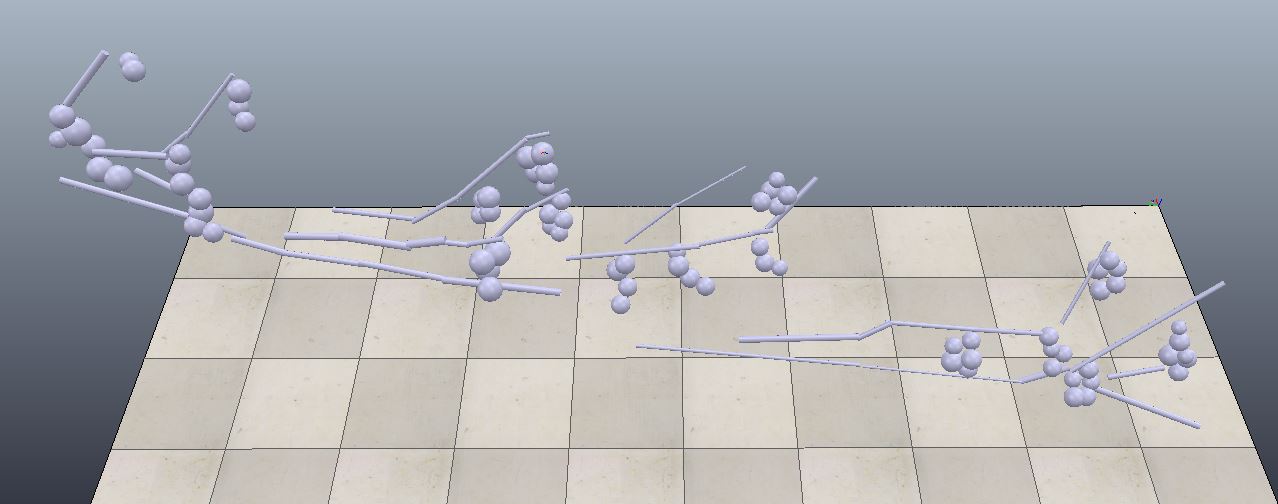}
	\caption{Simulation Environment}
	\label{fig:environment}
\end{figure}

With the mesh created, it was then imported into CoppeliaSim and ready to be converted into simple shapes that would represent the branches and tomatoes present on the vine. Once imported, cylinders were extracted from the mesh to represent branches and spheres were extracted to represent tomatoes. The provided greenhouse photos were used as reference to ensure the amount of tomatoes per bunch was accurate to reality. Using these images, the proportional sizes of the tomatoes was maintained. The branches were also sized using the provided set of images. Once all the tomatoes and environment obstacles have been created as simple shapes, the mesh was then able to be used with a simulated robotic arm. Figure \ref{fig:environment} shows the entirety of the simulated environment.

\subsection{Simulation of a robotics operation}

\begin{figure}[h]
	\includegraphics[width=0.5\textwidth]{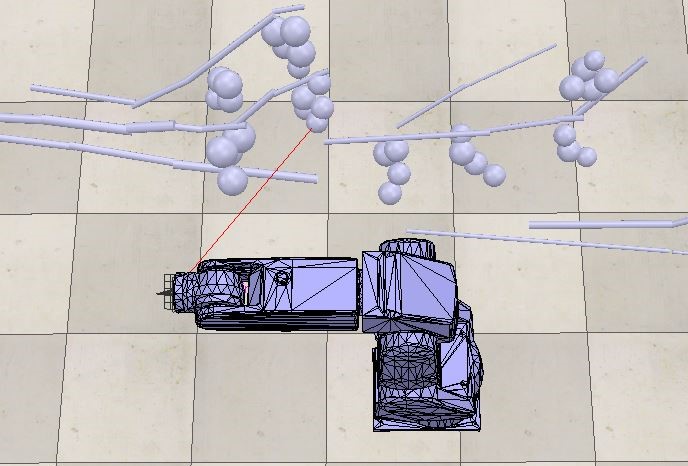}
	\caption{Robot Initial Position}
	\label{fig:initialPos}
\end{figure}

The goal of the simulation is to move the robot's end effector from its initial position to the position of the grasping target without making contact with any obstacles on the way. The grasping target can be changed by simple moving the position of the target in the simulation. Before the simulation begins working, two things have to be set for the simulator to continue. The first step is to create handles for every object within the simulation. A handle is a string that will refer to a respective object within the simulation. A handle must be obtained for every part of the robot and anything else that is manipulated in the simulation. After this, the robot is set to its initial position before each trial. This initial position can be easily changed if required, but currently the joint values are all set to 0, which corresponds to an initial configuration as seen in Figure \ref{fig:initialPos}. The red line is a straight line from the tip to the grasping target.

At this point, information regarding the target's position is retrieved. A randomized forward kinematics search for manipulator configurations which fall within a specified threshold of the end effector's position is conducted. If any configurations found are very similar to previously found configurations, the solution is discarded. Similarity is determined by whether or not each joint value is within 0.01 radians of the respective joint value in previously found solutions. From the randomized search, many configurations will be too far from the target to successfully compute the inverse kinematics, which is why a threshold is needed. This threshold is the maximum distance between tip and target indicating when inverse kinematics should attempt to bring the tip to the target. The larger the threshold, the longer the computation time. Conversely, if the threshold is too small, the subset of solutions will also be very small.  The program recommends a value of 0.65, but the optimal value is going to be influenced by the robot and the environment.

With an appropriate value selected for the threshold, the simulation is able to determine a set of acceptable final configurations. The planner will have 60 different final configurations from this step to choose from. This step is where the path planning algorithms are used, and where the majority of the computation time is consumed. The planner used at this step is user-defined; there are 25 different planners available in the OMPL library. At this point, the planner is told what the initial and final configurations are, and collision pairs are defined within the simulation, which defines obstacles and tells the planner what needs to be avoided in the simulation. It is at this point the path planning algorithm computes the path to move the robot's end effector from the initial position to the target position. Figure \ref{fig:finalPos} below shows the robot after the path planning is completed and the robot is at its target configuration. The pink line represents the path the end effector took to reach the final configuration.

\begin{figure}[h]
	\includegraphics[width=0.5\textwidth]{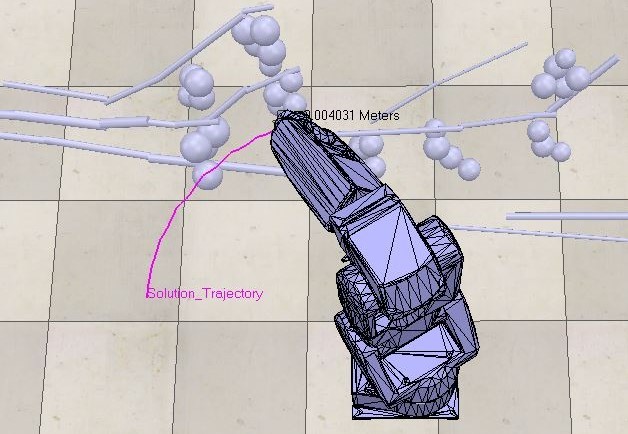}
	\caption{Robot Final Position}
	\label{fig:finalPos}
\end{figure}

\section{Discussion}

From the method shown in this paper, we can create a realistic greenhouse environment that captures some the real world greenhouse complexity. Regions in the image set which are greater than 60 centimeters from the camera are discarded, removing background noise and allowing path planning algorithms to focus on avoiding obstacles that are actually problematic for the grasp plan. Grasp planning and obstacle avoidance can be studied and improved through the use of real greenhouse data.  The environment created from the greenhouse data naturally contains tomatoes which are easy to grasp, some that are hard, and some that are impossible without manipulating the surrounding clutter. This has resulted in a much more realistic complexity distribution than manually placed tomatoes and obstacles. 

\subsection{Future Work}

Future improvement of this work will focus on extracting more complex shapes from the pointcloud to more closely represent objects in the environment rather than cylinders for vines and spheres for tomatoes. To achieve this, the pointcloud would require more information such as a side or rear view. However, this would be very difficult due to the clutter levels of each row and implementing a rear view for every front view in the greenhouse would be very difficult and time consuming. Other improvement will focus on automating the process of capturing images to enable temporal simulation of the greenhouse development which will enable testing of robotics operation in the entire season taking into consideration various stages of plan growth. 

\section{Conclusion}
This paper has presented a method for creating realistic simulations of real vegetable greenhouse environments. These simulated environments can be used for testing robotics operations in these greenhouses such as harvesting. Doing this could greatly increase the harvest yield, resulting in monetary benefit for the greenhouse.

\end{document}